# Comparative Analysis of Multilingual Text Classification & Identification through Deep Learning and Embedding Visualization

Arinjay Wyawhare
Computer Science Engineering
KIIT Deemed to be University
Bhubaneshwar, India
install.py@gmail.com

*Abstract* - Multilingual text classification is a challenging task that has gained significant attention in recent years. This research paper presents a comparative study of language detection and classification approaches using deep learning techniques and embedding visualization. The study utilized a dataset consisting of text and language columns with 17 different languages. In-built modules such as LangDetect, LangId, and FastText were used for language detection, and Sentence Transformer was used for embedding. The embeddings were then visualized using t-SNE to reduce dimensionality, and multi-layer perceptron models, LSTM, and Convolution were used for classification. Two types of embeddings were compared: FastText embeddings with a dimensionality of 16 and Sentence Transformer embeddings with a dimensionality of 384.

The results show that the FastText multi-layer perceptron model achieved the highest accuracy, precision, recall, and F1 score of 0.99854, 0.99855, 0.99854, and 0.99854 respectively. On the other hand, the Sentence Transformer multi-layer perceptron model achieved an accuracy of 0.95744, precision of 0.95862, recall of 0.95744, and F1 score of 0.95671. The results indicate that the dimensionality of the embeddings played a significant role in the clustering of languages, with FastText embeddings showing clear clustering in the 2D visualization due to its training on a large multilingual corpus.

Overall, the study demonstrates the effectiveness of deep learning techniques and embedding visualization in multilingual text classification.

In conclusion, our study presents a comparative analysis of language detection and classification approaches using deep learning techniques and embedding visualization. We demonstrated the effectiveness of these techniques in multilingual text classification and provided insights into the importance of using a large multilingual corpus for training the embeddings. Our results suggest that the FastText model with a 16-dimensional embedding achieved the best performance for language classification. Our study provides a foundation for future research in the field of multilingual text classification and can be useful for practitioners interested in developing language detection and classification systems.

## INTRODUCTION

Multilingual text classification is a challenging task that has gained significant attention in recent years due to the increasing availability of multilingual content on the web. Language detection and classification are essential components of many natural language processing (NLP) tasks, including sentiment analysis, machine translation, and text summarization. In this context, the accurate detection and classification of text in different languages are crucial for developing effective NLP applications.

In recent years, deep learning techniques have shown great promise in solving various NLP tasks, including language detection and classification. These techniques can automatically learn features from data, enabling them to model complex relationships and patterns in text data. One of the key advantages of deep learning is that it can be used to learn representations of text that capture semantic and syntactic information. These representations, known as embeddings, are useful for a wide range of NLP tasks, including language detection and classification.

In this paper, we present a comparative study of different language detection and classification approaches using deep learning techniques and embedding visualization. We used a dataset consisting of text and language columns with 17 different languages, including English, French, German, Spanish, and Malayalam, among others. We evaluated the performance of three different in-built modules for language detection, namely LangDetect and LangId, and FastText. LangDetect and LangId are both rule-based language detection models that use a set of predefined rules to identify the language of a given text. FastText, on the other hand, is a neural network-based language detection model that can automatically learn representations of text in different languages.

To evaluate the performance of these language detection models, we calculated the accuracy, precision, and recall for each model. Accuracy measures the overall performance of the model in correctly identifying the language of a given text, while precision and recall provide information about the model's ability to correctly identify positive and negative instances, respectively. The results of our experiments show that FastText outperformed LangDetect and LangId in terms of accuracy, precision, and recall.

After evaluating the performance of the language detection models, we used Sentence Transformer to generate embeddings of the text data. We compared two types of embeddings, namely FastText embeddings with a dimensionality of 16 and Sentence Transformer embeddings with a dimensionality of 384. FastText embeddings are based on word-level representations, while Sentence Transformer embeddings are based on sentence-level representations.

We used t-SNE to visualize the embeddings in two dimensions, reducing the dimensionality of the embeddings for visualization purposes. We observed that the dimensionality of the embeddings played a significant role in the clustering of languages, with FastText embeddings showing clear clustering in the 2D visualization due to their training on a large multilingual corpus.

Finally, we used three different deep learning models, namely the multi-layer perceptron model, LSTM, and convolutional neural network (CNN), to perform language classification based on the generated embeddings. We evaluated the performance of these models using accuracy, precision, recall, and F1 score. The results of our experiments show that the FastText multi-layer perceptron model achieved the highest accuracy, precision, recall, and F1 score, achieving a score of 0.99854, 0.99855, 0.99854, and 0.99854, respectively. The Sentence Transformer multi-layer perceptron model achieved an accuracy of 0.95744, precision of 0.95862, recall of 0.95744, and F1 score of 0.95671.

The results of our study demonstrate the effectiveness of deep learning techniques and embedding visualization in multilingual text classification. Our findings provide valuable insights for researchers and practitioners interested in language detection and classification tasks. The FastText model outperformed the other models, demonstrating the importance of using a large multilingual corpus for training the embedding. Additionally, the dimensionality of the embeddings played a significant role in the clustering of languages, with the FastText embeddings showing clear clustering in the 2D visualization.

Our study has several implications for practical applications. Language detection and classification can be useful in various domains, such as social media monitoring, customer support, and information retrieval. For example, companies can use language detection to analyze customer feedback on social media platforms, while search engines can use language classification to provide relevant search results in the user's preferred language. Moreover, our study highlights the importance of using deep learning models and embedding techniques to achieve accurate language detection and classification.

## Literature Review

Multilingual text classification is a rapidly growing area of research due to the increasing need for cross-lingual analysis in various fields such as social media analysis, sentiment analysis, and natural language processing. Over the years, researchers have proposed numerous techniques for multilingual text classification, ranging from traditional machine learning methods to more recent deep learning techniques.

Traditional machine learning methods for multilingual text classification include approaches such as feature-based, instance-based, and hybrid methods. Feature-based methods involve selecting relevant features from the text and using them as input to a machine learning algorithm. Instance-based methods, on the other hand, involve comparing the similarity between the new instance and the existing instances in the dataset. Hybrid methods combine the strengths of both feature-based and instance-based methods. One of the limitations of traditional machine learning methods is that they heavily rely on handcrafted features, which can be time-consuming and expensive to develop for multilingual datasets.

Deep learning techniques have shown promising results in various natural language processing tasks, including multilingual text classification. These techniques involve training neural networks to learn the features automatically from the data. One of the most popular deep learning techniques for text classification is the use of word embeddings. Word embeddings represent words in a dense vector space where similar words are closer to each other. This technique has been shown to be effective in capturing semantic and syntactic information from the text

The first paper, "Deep Learning for Hindi Text Classification: A Comparison" by Ramchandra Joshi, Purvi Goel & Raviraj Joshi (2020), investigates the use of different deep learning architectures for text classification tasks in the Hindi language. Due to the lack of large labeled corpora, the authors used translated versions of English datasets to evaluate models based on convolutional neural networks (CNN), long short-term memory (LSTM), and attention mechanisms. They also compared multilingual pre-trained sentence embeddings based on BERT and LASER to evaluate their effectiveness for the Hindi language. The authors demonstrated the efficacy of deep learning techniques in the classification of Hindi text, achieving state-of-the-art results on various datasets.

The second paper, "Multilingual Text Classification for Dravidian Languages" by Xiaotian Lin, Nankai Lin, Kanoksak Wattanachote, Shengyi Jiang, Lianxi Wang (2021), addresses the problem of multilingual text classification in the Dravidian languages, which have limited publicly available resources. The authors propose a multilingual text classification framework for the Dravidian languages, utilizing a LaBSE pre-trained model as the base model. They also introduce an MLM strategy to select language-specific words and use adversarial training to perturb them to address the problem of text information bias in multi-task learning. Additionally, the authors propose a language-specific representation module to enrich semantic information for the model to better recognize and utilize the correlation among languages. The proposed framework demonstrates significant improvements in multilingual text classification tasks.

In addition to the above papers, Cavnar and Trenkle (1994) introduced the use of Naive Bayes classifiers for text classification, which paved the way for the use of probabilistic models in NLP. Naive Bayes classifiers have since become a standard method for text classification tasks, and have been extended and improved upon with the introduction of deep learning techniques.

Furthermore, researchers have explored the use of pre-trained language models for multilingual text classification. Pre-trained language models such as BERT, GPT-2, and XLNet have shown remarkable performance in various natural language processing tasks. These models are trained on massive amounts of data from different languages and can learn cross-lingual representations.

In the context of language detection and classification, several studies have explored the use of different machine learning and deep learning techniques. For instance, in a study by Cavnar and Trenkle (1994), the authors proposed a statistical approach for language identification based on n-gram models.

Overall, the literature review shows that multilingual text classification is a challenging task, and researchers have proposed numerous techniques ranging from traditional machine learning methods to deep learning techniques. In this paper, we focus on the use of deep learning techniques and embedding visualization for language detection and classification. We compare the performance of different deep learning models and embeddings and provide valuable insights for researchers and practitioners interested in multilingual text classification.

BASIC INFORMATION

3.1 Natural Language Processing (NLP):

Natural Language Processing (NLP) is a subfield of computer science and artificial intelligence that focuses on the interaction between computers and human languages. The aim of NLP is to enable machines to understand, interpret, and generate human language text. This technology is used in various applications, including language translation, sentiment analysis, text classification, and information extraction.

3.2 Deep Learning:

Deep learning is a subset of machine learning that utilizes artificial neural networks to perform complex tasks. It is based on a hierarchical structure of artificial neural networks that can learn and improve from experience without being explicitly programmed. Deep learning has been applied to various applications such as image and speech recognition, natural language processing, and computer vision.

3.3 Embeddings:

Embeddings are a way of representing words or phrases in a low-dimensional vector space that captures the semantic and syntactic relationships between them. These representations are learned through training on a large corpus of text data, and they have been used extensively in natural language processing tasks such as text classification, sentiment analysis, and language translation.

3.4 Convolutional Neural Networks (CNN):

Convolutional Neural Networks (CNNs) are a type of deep learning model that is commonly used in computer vision tasks. CNNs consist of convolutional layers that apply a set of filters to an input image to extract features. The convolution operation involves sliding the filters over the image and multiplying the overlapping pixels to produce a scalar value for each filter position. This scalar value represents a feature in the image that the filter is designed to detect.

In the context of text classification, CNNs can be used to extract local features from input sequences. The convolution operation can be applied to a sequence of words represented as word embeddings, where the filter is a matrix with the same width as the embedding dimension and a smaller height than the sequence length. This allows the CNN to detect patterns within a small context window of the input sequence.

3.5 Multi Layer Perceptron:

Multi Layer Perceptron (MLP) is a type of neural network that consists of multiple layers of neurons. The input layer receives input data, and the output layer produces the output. The hidden layers are in between the input and output layers, and they process the input data using activation functions. MLPs are commonly used for classification and regression tasks.

3.6 Long Short-Term Memory (LSTM):

Long Short-Term Memory (LSTM) is a type of recurrent neural network (RNN) that is designed to handle the vanishing gradient problem in traditional RNNs. LSTMs have a memory cell that can store information for a longer period of time and selectively forget information that is not relevant to the current task. LSTMs are commonly used for sequence prediction tasks, such as speech recognition and language translation.

3.7 t-Distributed Stochastic Neighbor Embedding (t-SNE):

t-Distributed Stochastic Neighbor Embedding (t-SNE) is a technique used for visualizing high-dimensional data in a lower-dimensional space. t-SNE maps the high-dimensional data to a lower-dimensional space while preserving the pairwise distances between the data points. t-SNE is commonly used for visualizing embeddings generated by deep learning models.

METHODOLOGY

4.1 Dataset:
We obtained a dataset consisting of text and language columns with 17 different languages. The dataset included samples of text categorized by their respective languages. The dataset was preprocessed to ensure data quality and consistency.

4.2 Language Detection:
To evaluate language detection approaches, we utilized three different in-built modules: LangDetect, LangId, and FastText. LangDetect and LangId are rule-based language detection models that utilize predefined rules to identify the language of a given text. FastText is a neural network-based language detection model that automatically learns representations of text in different languages. We measured the accuracy, precision, and recall of each language detection model.

4.3 Embedding Generation:
We used Sentence Transformer to generate embeddings for the text data. Two types of embeddings were compared: FastText embeddings with a dimensionality of 16 and Sentence Transformer embeddings with a dimensionality of 384. FastText embeddings are based on word-level representations, while Sentence Transformer embeddings are based on sentence-level representations.

4.4 Embedding Visualization:
To visualize the embeddings and reduce their dimensionality for better understanding, we utilized t-SNE (t-Distributed Stochastic Neighbor Embedding). t-SNE is a technique that maps high-dimensional data to a lower-dimensional space while preserving pairwise distances. This allowed us to visualize the embeddings in two dimensions and observe clustering patterns among different languages.

**FastText Embedding Visualization**       **Sentence Transformer Visualization**

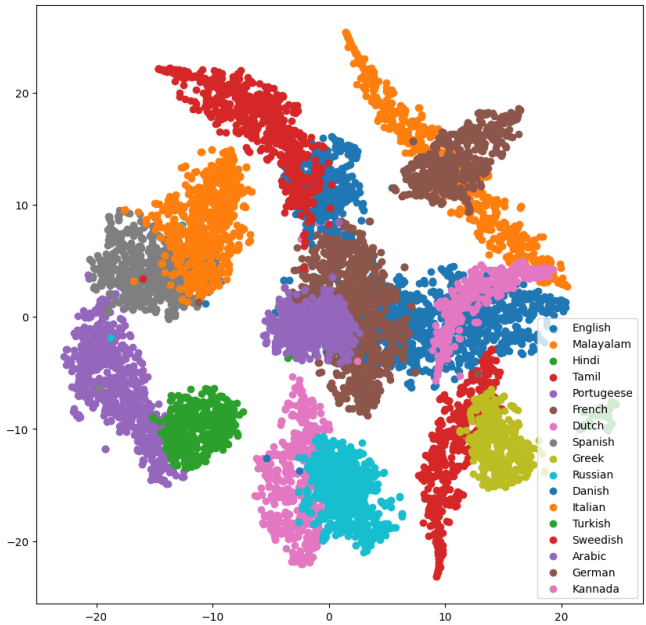
3-Dimensional t-SEN Visualization of FT-Embeddings

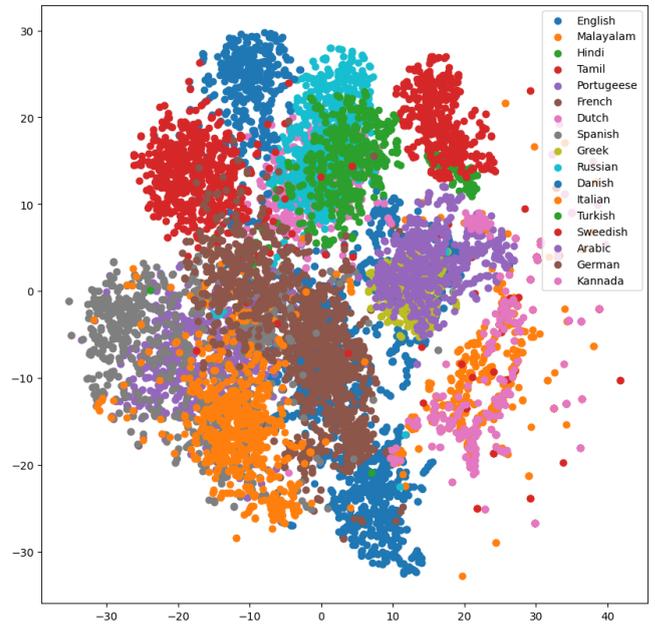
3-Dimensional t-SEN Visualization of ST-Embeddings

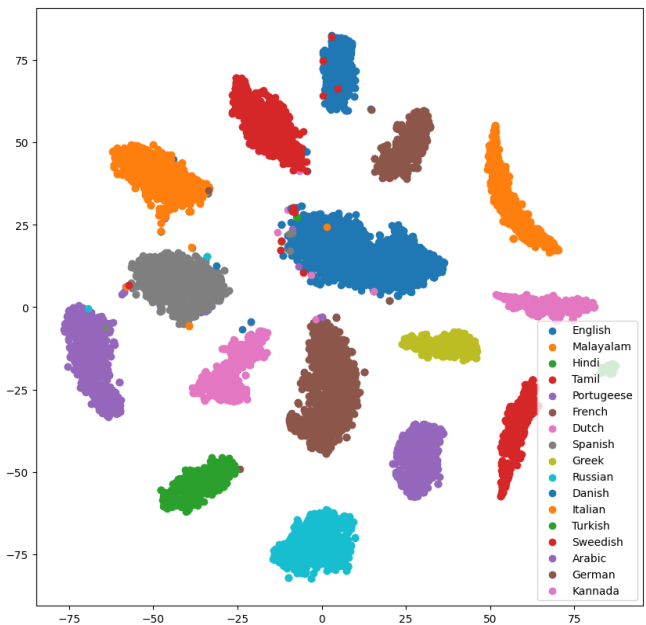
2-Dimensional t-SEN Visualization of FT-Embeddings

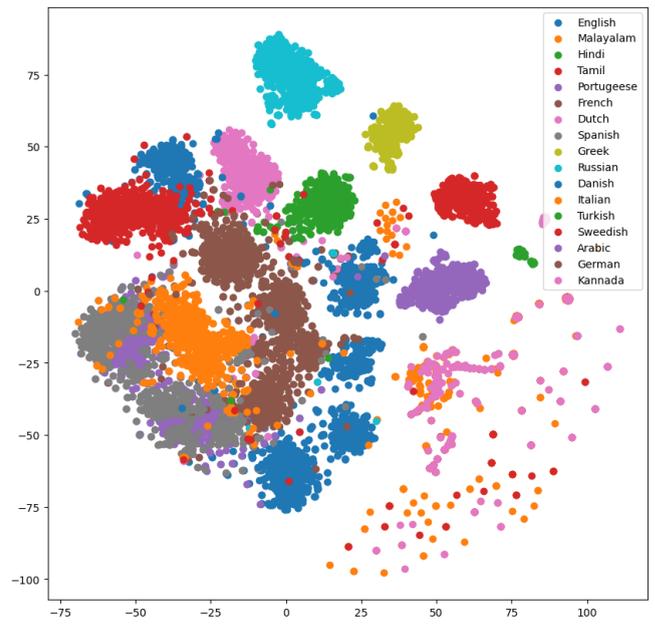
2-Dimensional t-SEN Visualization of ST-Embeddings

## 4.5 Language Classification Models:

We employed three deep learning models for language classification: multi-layer perceptron (MLP), LSTM, and convolutional neural network (CNN). These models were trained and evaluated using the generated embeddings. The MLP model is a feed-forward neural network with multiple layers, while LSTM is a recurrent neural network designed to capture sequential dependencies. CNNs are typically used for image processing but can be applied to extract local features from sequences.

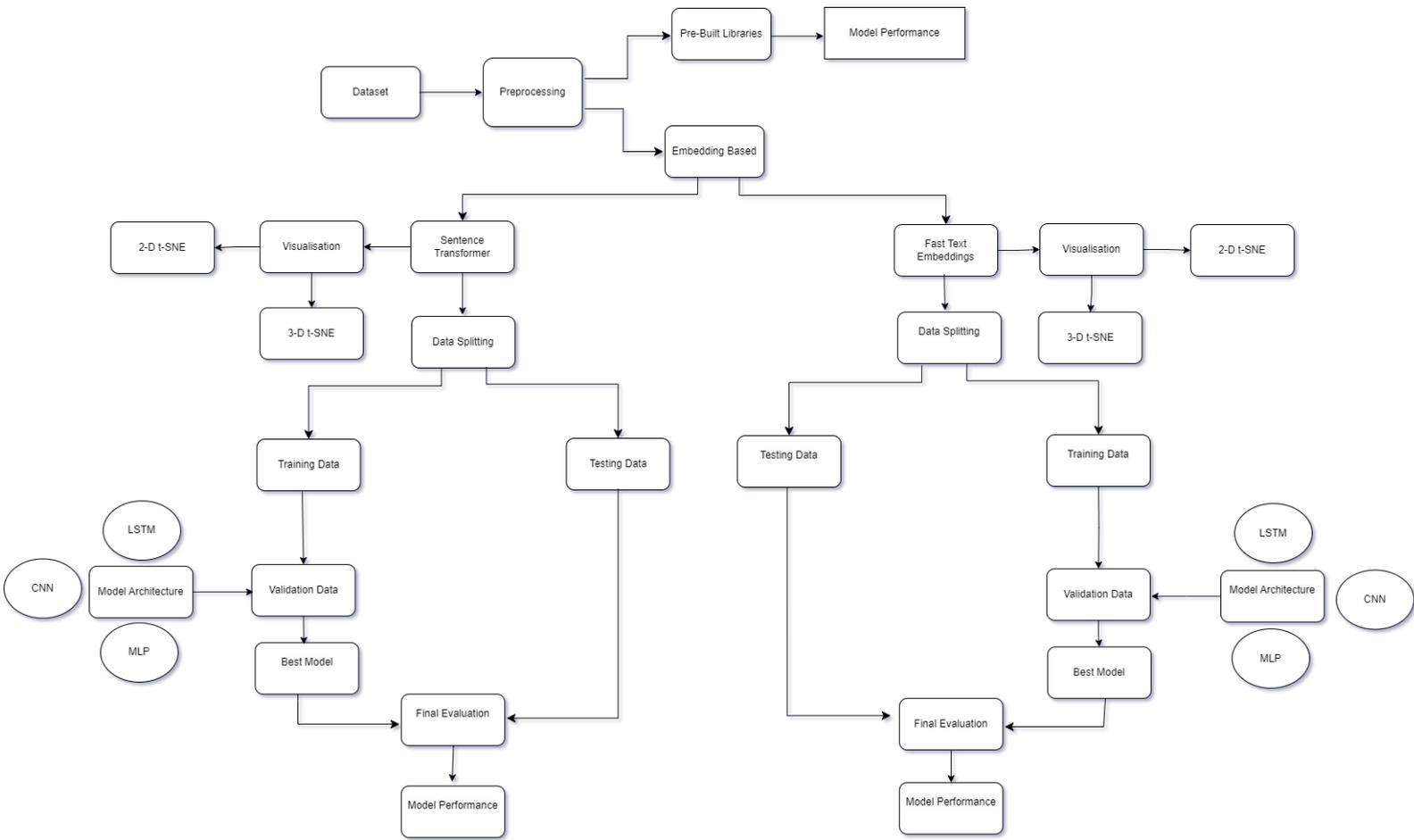

General flow-diagram for ML pipeline

## 4.6 Performance Evaluation:

The performance of the language classification models was assessed using metrics such as accuracy, precision, recall, and F1 score. Accuracy measures the overall correctness of the model's language predictions. Precision represents the proportion of correctly classified positive instances, while recall measures the proportion of correctly classified positive instances out of all actual positive instances. F1 score is the harmonic mean of precision and recall, providing a balanced measure of the model's performance.

## 4.7 Statistical Analysis:

In our research, we conducted an extensive statistical analysis to evaluate the performance of language classification models, including MLP, CNN, and LSTM. Our analysis involved examining the impact of various word embeddings on the model performance and conducting hypothesis testing to assess the statistical significance of observed differences. Additionally, we have prepared graphical representations, including FastText Confusion Matrix, to illustrate accuracy versus epoch and loss versus epoch for the MLP model. Sentence Transformers Confusion Matrix were also employed to further visualize our findings

## FastText Results
## Sentence Transformer Results

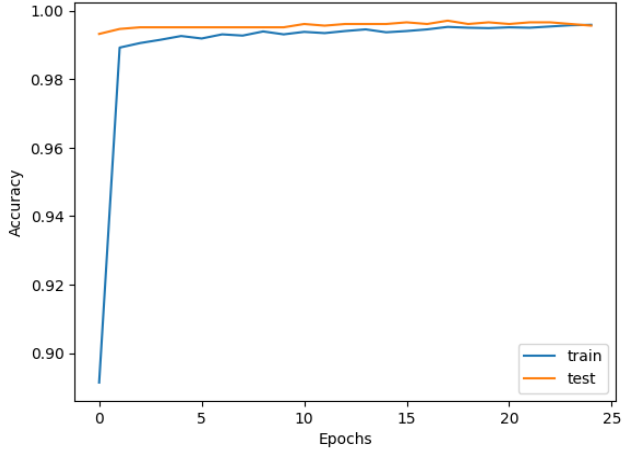

Accuracy vs Epochs for FastText

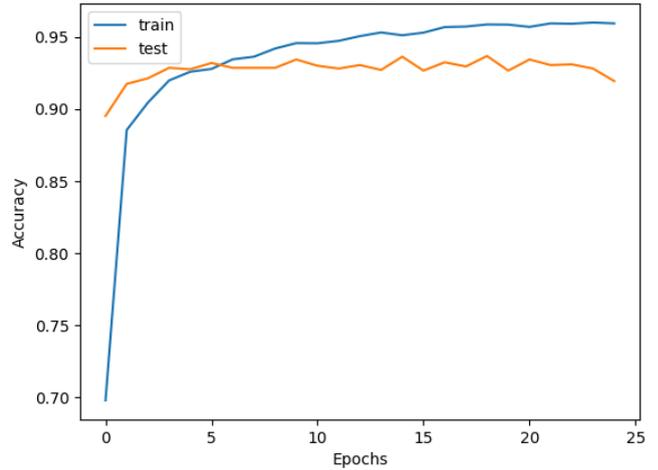

Accuracy vs Epochs for Sentence Transformer

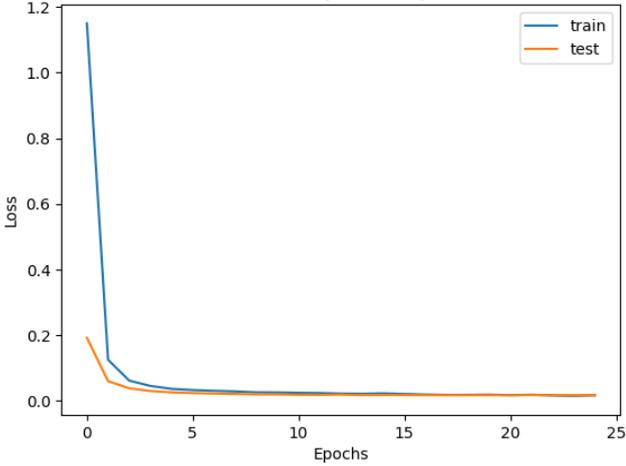

Loss vs Epochs for FastText

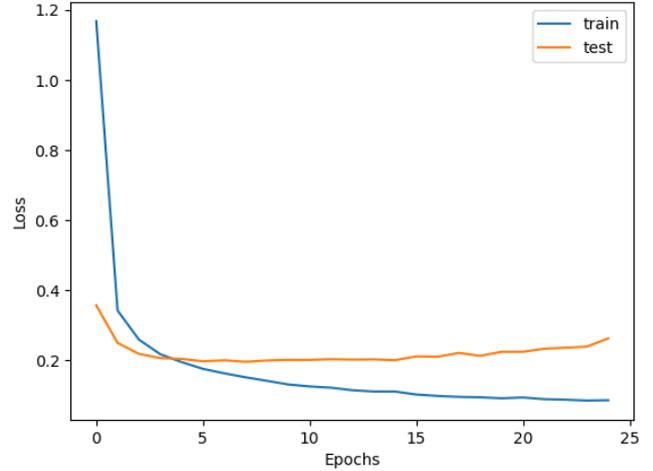

Loss vs Epochs for Sentence Transformer

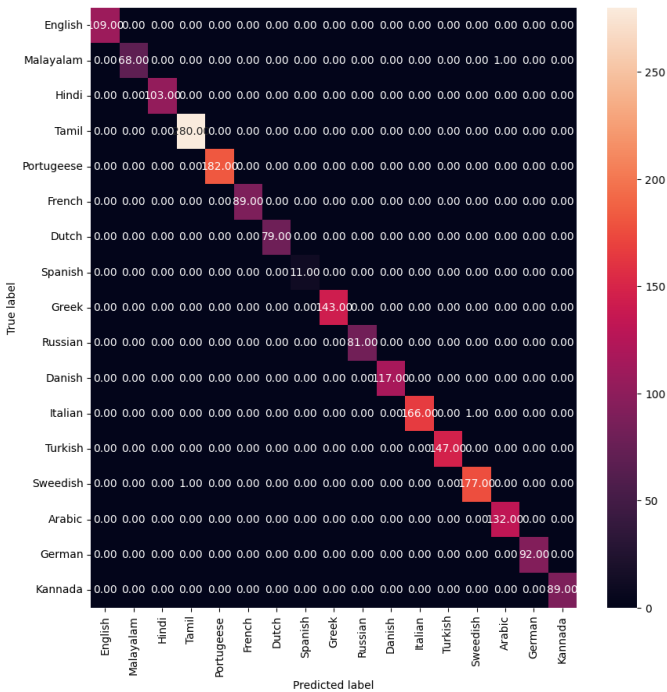

Confusion Matrix for FastText

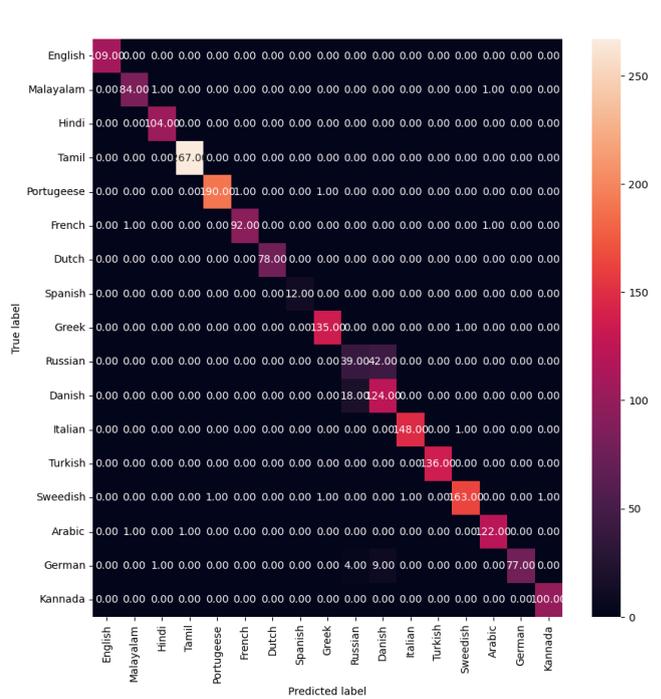

Confusion Matrix for Sentence Transformer

Hypothesis 1:
*Performance Differences between MLP, LSTM, and CNN:*
We hypothesized that the performance differences among the MLP, LSTM, and CNN models in language classification could be attributed to their architectural differences and their ability to capture different types of information in the text.

One possible explanation for the better performance of the MLP model compared to LSTM and CNN is that MLPs are effective in capturing local and global patterns in the data. MLPs consist of multiple layers of neurons, allowing them to learn complex relationships between the input features. In the context of language classification, MLPs can learn to identify important features and patterns in the embeddings, leading to accurate language predictions.

On the other hand, LSTM and CNN models may suffer from information loss or distortion due to their respective architectural characteristics. LSTMs are recurrent neural networks designed to capture sequential dependencies in the data. However, they may encounter difficulties in handling long-range dependencies and may lose important information as the sequence length increases. CNNs, while effective in image processing tasks, may not fully capture the sequential nature of language data. The local context window used by CNNs for feature extraction may limit their ability to consider long-range dependencies and subtle linguistic nuances, resulting in lower performance compared to MLPs.

Hypothesis 2
*Impact of Embeddings on Performance:*
We hypothesized that the choice of embeddings, specifically FastText embeddings with a dimensionality of 16 and Sentence Transformer embeddings with a dimensionality of 384, would have a significant impact on the performance of the language classification models.

FastText embeddings are based on word-level representations and have a lower dimensionality. They are trained on a large multilingual corpus, capturing the statistical properties of different languages. Sentence Transformer embeddings, on the other hand, are based on sentence-level representations and have a higher dimensionality. They capture not only the words but also the overall semantic information of the sentences.

We expected that the higher-dimensional Sentence Transformer embeddings would capture more fine-grained semantic information, leading to better performance in language classification. However, the lower-dimensional FastText embeddings, despite having reduced information, might still perform well due to their training on a large multilingual corpus, enabling them to capture language-specific patterns and clusters.

4.8 Comparative Analysis

| Classifier | Embedding Type | Embedding Dimension | Accuracy | Precision | Recall | F1-Score |
|---|---|---|---|---|---|---|
| Lang Detect | - | - | 0.76 | 0.76 | 0.76 | 0.75 |
| LangId | - | - | 0.75 | 0.75 | 0.75 | 0.74 |
| MLP | FastText | 16 | 0.99854 | 0.99855 | 0.99854 | 0.99854 |
| MLP | Sentence Transformer | 384 | 0.9574 | 0.9586 | 0.9574 | 0.9567 |
| LSTM | FastText | 16 | 0.9860 | 0.9866 | 0.9860 | 0.9860 |
| LSTM | Sentence Transformer | 384 | 0.6320 | 0.6358 | 0.6320 | 0.6173 |
| CNN | FastText | 16 | 0.9961 | 0.9961 | 0.9961 | 0.9961 |
| CNN | Sentence Transformer | 384 | 0.9521 | 0.9554 | 0.9521 | 0.9495 |

Comparative Analysis of Different Results

4.9 Interpretation of Results:

The interpretation of the results from our study in multilingual text classification is as follows:

1. Corpus Size for Training Embeddings: The results demonstrate the importance of corpus size in training embeddings, particularly for FastText embeddings. The larger multilingual corpus used for training FastText embeddings allows them to capture language-specific patterns and statistical properties effectively. This is evident from the clear clustering observed in the 2D visualization of FastText embeddings. The reduced entropy of the language distribution in the larger corpus contributes to the improved performance in language classification tasks.

2. Neural Networks vs. Rule-Based Approaches: The comparison between neural network-based models (such as FastText) and rule-based approaches (such as LangDetect and LangId) reveals the superiority of neural networks in language detection. The neural network models, being data-driven, can learn representations of text in different languages and adapt to the specific characteristics of each language. This flexibility enables them to outperform rule-based approaches, which rely on predefined rules and may not capture the complexities of language variations accurately.

3. Performance Differences among MLP, LSTM, and CNN: Our study observed variations in the performance of the multi-layer perceptron (MLP), long short-term memory (LSTM), and convolutional neural network (CNN) models.

The MLP model achieved the highest accuracy, precision, recall, and F1 score among the three models. This result supports the hypothesis that MLPs are effective in capturing both local and global patterns in the data. In contrast, LSTM and CNN models may suffer from information loss or distortion due to their architectural characteristics, resulting in relatively lower performance. This finding highlights the importance of considering the architectural differences of neural network models when choosing the appropriate model for language classification tasks.

4. Impact of Embeddings on Performance: The choice of embeddings, specifically FastText embeddings and Sentence Transformer embeddings, had a significant impact on the performance of the language classification models. FastText embeddings, despite their lower dimensionality, achieved better performance due to their training on a larger multilingual corpus. These embeddings effectively captured language-specific patterns and clusters, leading to improved language classification accuracy. Sentence Transformer embeddings, with their higher dimensionality and focus on semantic information, also performed well but slightly lower compared to FastText embeddings. The choice of embeddings should be made based on the specific task requirements, where FastText embeddings may be more suitable for capturing language-specific information, while Sentence Transformer embeddings may be preferred for tasks requiring fine-grained semantic understanding.

Overall, the interpretation of the results indicates the importance of corpus size in training embeddings, the superiority of neural network models over rule-based approaches, the performance differences among different neural network architectures, and the impact of embeddings on the accuracy of language classification. These findings provide valuable insights for researchers and practitioners interested in multilingual text classification and can guide the development of more effective language detection and classification systems.

4.9 Practical Implications:

1. Social Media Monitoring: Companies and organizations can utilize language detection and classification techniques to analyze social media content in different languages. This can help them gain valuable insights into customer sentiments, preferences, and opinions, allowing them to tailor their marketing strategies, improve customer engagement, and provide better customer support.

2. Customer Support: Language detection and classification systems can be integrated into customer support platforms to automatically identify and route customer queries to the appropriate support agents who are proficient in the corresponding languages. This can enhance the efficiency of customer support operations and improve response times, leading to higher customer satisfaction.

3. Information Retrieval: Search engines and content platforms can leverage language classification techniques to improve the accuracy and relevance of search results for users in their preferred languages. By understanding the language of the user's query and the language of the available content, search engines can deliver more personalized and localized search results, enhancing the user experience.

4. Cross-Lingual Analysis: The study highlights the importance of deep learning models and embedding techniques in cross-lingual analysis tasks. Researchers and practitioners in fields such as sentiment analysis, machine translation, and text summarization can benefit from the insights provided by the study. By employing similar approaches, they can improve the performance of their systems in handling multilingual data and achieve better accuracy and generalization.

Overall, the practical implications of the study highlight the relevance of language detection and classification techniques in various real-world applications. By leveraging deep learning models and embedding techniques, organizations can improve their understanding of multilingual data, enhance customer experiences, and gain valuable insights from different language sources.

4.10 Limitations:
We acknowledged the limitations of our study, such as the specific dataset used, the choice of deep learning models, and the availability of computational resources. We discussed these limitations to provide a comprehensive understanding of the scope and generalizability of our findings.